\documentclass[lettersize,journal]{IEEEtran}
\usepackage{amsmath,amsfonts}
\usepackage{algorithmic}
\usepackage{algorithm}
\usepackage{array}
\usepackage[caption=false,font=normalsize,labelfont=sf,textfont=sf]{subfig}
\usepackage{textcomp}
\usepackage{stfloats}
\usepackage{url}
\usepackage{verbatim}
\usepackage{graphicx}
\usepackage{cite}
\usepackage{multirow}
\usepackage{rotating}
\usepackage[super]{nth}

\newcommand{\etal}{\textit{et al.~}}

\begin{document}

\title{Learning Long-Term Temporal Dependencies in Photovoltaic Power Output Prediction Through Multi-Horizon Forecasting}

\author{
\IEEEauthorblockN{Sumit Laha, Ankit Sharma, Hassan Foroosh}
\\
\IEEEauthorblockA{Department of Computer Science\\
University of Central Florida\\
Orlando, Florida, United States.\\
sumit.laha@ucf.edu, ankit.sharma@ucf.edu, hassan.foroosh@ucf.edu}
}



\maketitle

\begin{abstract}
The rapid global expansion of solar photovoltaic (PV) capacity—reaching a record 597 GW in 2024—highlights the urgent need for robust forecasting models to mitigate the grid instability caused by the intermittent nature of solar irradiance. While deep learning-based direct forecasting using ground-based sky images (GSI) has emerged as a dominant approach, existing literature is often constrained by single-architecture evaluations and an exclusive focus on single-horizon (point) prediction. This paper proposes a transition from traditional single-horizon estimation toward a multi-horizon forecasting framework, leading to an architecture-independent improvement in accuracy. We hypothesize and demonstrate experimentally that joint optimization over a sequence of future values allows deep neural networks to better capture latent inter-step temporal dependencies by avoiding precocious convergence of the network in terms of both weight gradients and filter diversity. Leveraging this architecture-independent improvement that integrates sequential sky imagery with historical PV generation data, we evaluate the models' abilities to predict power output across multiple discrete future time steps simultaneously. Our methodology is validated through a comparative analysis across diverse deep learning architectures. The results demonstrate that this multi-horizon approach significantly enhances predictive accuracy and robustness across the entire forecast horizon while maintaining computational parsimony. By achieving superior performance with negligible overhead compared to single-horizon models, this work provides a scalable and efficient solution to improve the resilience of modern power grids.
\end{abstract}


\section{Introduction}
\IEEEPARstart{D}{uring} the past decade, there has been considerable interest in deploying solar photovoltaic (PV) grids as a crucial component of the transition to a sustainable and cleaner energy system \cite{gielen2019role,kabir2018solar}. In 2024, a record 597 GW of new solar capacity was installed, representing a 33\% increase over the previous year. This surge reinforces the leading role of solar energy in the transition to a cleaner energy system\footnote{https://www.solarpowereurope.org/press-releases/new-report-world-installed-600-gw-of-solar-in-2024-could-be-installing-1-tw-per-year-by-2030}. However, the unpredictability of solar power generation presents a significant challenge to its widespread deployment and economic operation. This variability is primarily attributed to two factors: seasonal variations in solar irradiance and sudden, unforeseen weather changes \cite{barbieri2017very,sun2019short}. Under cloudy conditions, PV output can fluctuate drastically on a minute-to-minute basis. Such rapid changes can negatively affect the stability and reliability of the electrical grid. Consequently, the development of accurate prediction of PV models is a critical step toward ensuring grid resilience and unlocking the full potential of renewable energy.
\IEEEpubidadjcol
Based on previous research, sky image-based short-term solar forecasting methods can be broadly categorized into two main approaches: indirect forecasting and direct forecasting \cite{RUAN}. Indirect methods extract cloud information, such as red-to-blue ratio \cite{long2006retrieving,ghonima2012method}, cloud coverage, and cloud motion vector from ground-based sky images (GSI) \cite{willert1991digital,beauchemin1995computation}. This approach employs a two-stage process: First, salient features are extracted from the input sky images, and then these features are used as inputs for either physical deterministic models \cite{chow2011intra,marquez2013intra,quesada2014cloud} or data-driven models, such as artificial neural networks \cite{chu2013hybrid,chu2015real,chu2015short,pedro2019adaptive}, to generate predictions. In contrast, direct forecasting approaches learn an end-to-end mapping from ground-based sky images to the target output (solar irradiance or PV power). This is accomplished by leveraging deep learning models, which have demonstrated superior performance due to their robust feature extraction capabilities and ability to learn from large datasets without the need for complex preprocessing steps. Consequently, deep learning-based direct forecasting methods are gaining prominence due to their capacity to autonomously identify predictive patterns in raw image data. This eliminates the dependency on handcrafted features and positions them as a robust and scalable solution for high-accuracy solar forecasting.

Deep learning has shown considerable promise in solar forecasting, exhibiting superior performance over traditional methods. However, the current body of research is subject to notable limitations that hinder the full realization of its potential in this domain. A key limitation of existing research is the use of relatively small datasets for model training and validation, with some studies relying on data collected over only a few weeks to months. This scarcity of data hinders the generalization capabilities of deep learning models, limiting their robustness when applied to unseen conditions. Furthermore, most research reports are based on only one deep network architecture, such as a single CNN variant \cite{sun2019short,nie2020pv,feng2020solarnet,paletta2020convolutional,nie2021resampling,feng2022convolutional}. This narrow focus limits the ability to draw general conclusions about a model's performance. Therefore, a thorough evaluation should include a comparative analysis across a variety of architectures to ensure the method's effectiveness is not tied to a specific network design. 

Accurate forecasting of solar irradiance and PV power output is critical to stable and efficient integration of solar energy into modern power grids. The prevailing methodology in the literature frames this challenge as a single-step or single-horizon forecasting problem. In this paradigm, models are trained to predict a single value of solar irradiance or PV power for one specific, pre-defined point in the future. Although this approach has demonstrated satisfactory performance and has become a standard practice, it may not fully exploit the rich temporal dependencies inherent in solar data. This leads us to a fundamental research question: Can a deep learning model, when trained to forecast multiple future time points simultaneously, more effectively capture the intricate temporal dependencies that govern the evolution of solar conditions?

In this paper, we propose a new methodology for predicting the photovoltaic power output in discrete future time steps. We hypothesize that joint optimization over a sequence of future values enables the learning of latent inter-step dependencies, thereby enhancing predictive ability across the entire forecast horizon. This investigation seeks to transition the solar forecasting paradigm from traditional single-point estimation toward a holistic, multi-horizon framework. Specifically, our proposed method uses a multimodal input stream comprising sequential sky imagery and historical PV generation data to generate predictions at multiple subsequent intervals. This design is in contrast to existing methodologies that are predominantly limited to single-step estimation. Furthermore, our results demonstrate that this joint multi-horizon approach achieves superior predictive accuracy while maintaining computational parsimony, incurring only negligible overhead compared to traditional single-output models.

The novelty and main contributions of this work are summarized as follows: 

\begin{itemize}
    \item We demonstrate that joint multi-horizon optimization enables learning long-term temporal dependencies in PV output forecasting.
    \item The concept is architecture-independent and is therefore proposed as an approach to improve existing forecasting models, rather than superseding them.
    \item Experimental results demonstrate that this idea yields superior accuracy with minimal increment in computational cost, making it highly suitable for real-time grid integration.
\end{itemize}

\section{Related work}
\label{sec_review}

Previous efforts in solar forecasting are generally categorized as indirect or direct methods \cite{RUAN}. This section provides a detailed review of the most relevant work in these two categories.

\subsection{Indirect Forecasting}
A key study by Marquez \cite{MARQUEZ} details an image processing methodology utilizing Total Sky Imagers (TSIs) to generate short-term forecasts of Direct Normal Irradiance (DNI) at ground level. Their work specifically describes relevant techniques for solar forecasting, including velocity field calculations, spatial image transformations, and cloud classification algorithms. \cite{CHU-indirect} developed a "smart" DNI forecasting model utilizing Artificial Neural Networks (ANNs). This model's unique strength lies in its fusion of cloud coverage time-series (extracted from a Total Sky Imager) with high-quality historical DNI time-series to generate predictions for short horizons of 5 and 10 minutes. Fu explored using image-based features to predict solar irradiance by first identifying a subset highly correlated with the desired target \cite{FU-indirect}. A regression model was trained using these selected features, with a clearness index conversion technique incorporated into the mechanism to improve prediction accuracy for the 5- to 15-minute horizon. 

\subsection{Direct Forecasting}

Direct forecasting models establish an end-to-end mapping that relates input ground-based sky imagery and historical data directly to the desired output, such as solar irradiance or PV power. A key appeal of this single-step learning paradigm is its ability to circumvent the limitations of multi-stage methods, providing specific advantages in model efficiency and robustness:

\begin{itemize}
    \item Bypassing Feature Engineering: Eliminating the need for laborious, multi-stage preprocessing, such as cloud detection, cloud tracking, and the calculation of physical parameters (e.g., optical flow, cloud base height).

    \item Reduced Cascading Error: By collapsing multiple intermediate stages into a single function, the method minimizes the accumulation and propagation of errors inherent in multi-step approaches.
\end{itemize}

\begin{figure*}[!t]
\centering
\includegraphics[scale=0.9]{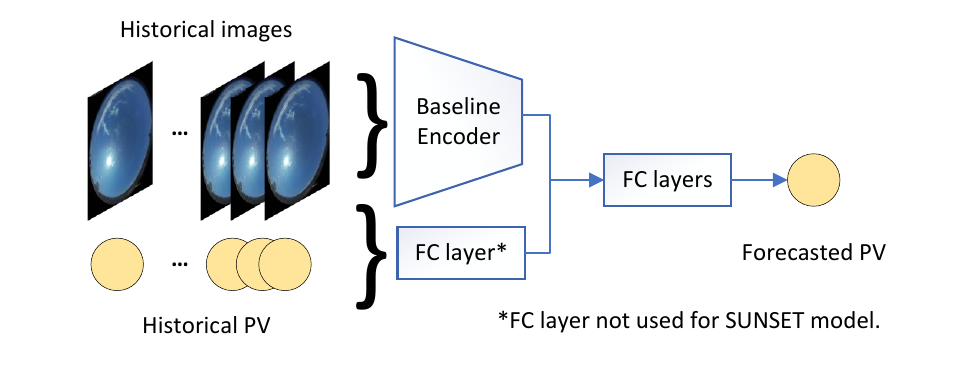}
\caption{Model-independent PV power generation framework based on both the single-point and the multi-horizon prediction tasks. The model takes sky images and PV power outputs from the past 15 minutes at an interval of 1 minute to predict future PV output. The single-point prediction forecasts a single PV output at ($t+T$) minutes into the future, while the multi-horizon prediction forecasts multiple outputs from time ($t+1$) to ($t+T$) minutes into the future at an interval of 1 minute.}\label{fig:framework}
\end{figure*}

\subsubsection{Shallow Networks}
One of the initial investigations into solar forecasting employed a basic Multilayer Perceptron (MLP) architecture \cite{MELLIT}. This simple approach utilized easily accessible features — mean solar irradiance, air temperature, and the day of the month—to predict solar irradiance with a 24-hour look-ahead horizon. Rana \cite{RANA} explored two distinct modeling strategies: univariate models, which rely solely on historical PV power data, and multivariate models, which incorporate auxiliary meteorological data alongside previous power outputs. Taravat \cite{Taravat} conducted a comparative analysis to assess the performance of an MLP against Support Vector Machines (SVM) for the task of cloud classification. Sahin \cite{SAHIN} proposed a method that integrated both the physical parameters of the PV panel and relevant weather data to develop a model. This model was designed to predict both the power production from the PV plant and the efficiency of the power plant. 

Leelaruji proposed a hybrid methodology for generating advanced warning (trigger) events for imminent sun coverage, achieving a prediction horizon of 1 to 2 minutes \cite{leelaruji2020short}. Their system combines classic image processing techniques—including the Hough transform for sun position and optical flow for cloud tracking—with a ResNet-based CNN to generate the final prediction. Zuo developed a deep learning-based model for the prediction of 10-minute-ahead GHI, designed to bypass errors associated with rigid cloud motion assumptions \cite{ZUO-indirect}. Their novel approach utilizes a hybrid cloud detection method along with a sophisticated set of inputs—including relative humidity, Aerosol Optical Depth (AOD), and prior GHI values—which are processed by a Long Short-Term Memory (LSTM) network optimized via Bayesian Optimization (BO). 

\subsubsection{Deep Neural Networks}
The work of Sun \cite{sun2018solar} marked a significant milestone, introducing a specialized CNN architecture named SUNSET to achieve the first direct image-based PV power nowcast from ground-based sky imagery. Later, they built on this foundation, adapting the model to successfully forecast PV output for a 15-minute look-ahead horizon \cite{SUN-Solar}. The SolarNet model \cite{FENG-Solarnet} predicted global horizontal irradiance (GHI) exclusively using only sky-images. This approach minimized complexity by eliminating the reliance on weather data and traditional feature engineering processes. Zang \cite{ZANG} proposed a two-stream network to independently extract spatial and temporal features, which are then adaptively fused using a gate unit. Following this fusion, a PV-guided attention mechanism is introduced to highlight dominant predictive regions before the features are fed into a progressive time series architecture for final PV power forecasting. 

In \cite{JONATHAN}, the authors proposed a method to integrate the attention mechanism into the CNN architecture, allowing the model to adaptively focus on the most relevant features of the sky image for a better prediction of GHI, DNI, and Diffuse Horizontal Irradiance (DHI). Lu \cite{lu2025enhanced} proposed CL-SUNSET, a novel framework for 15-minute-ahead PV power forecasting that integrates self-supervised contrastive learning with a CNN-based regression model. This integration facilitates the effective utilization of information from large volumes of unlabeled ground-based sky images. Liu \cite{LIU} introduced the Spatial–Temporal Multimodal Fusion Model (STMFM) model, which is a novel dual-stream feature extraction structure designed to isolate spatial and temporal information. This architecture processes the original image and power sequences in one stream (spatial relationships) and the optical flow and power difference sequences in the other (temporal evolution), enabling a more effective extraction of specialized features. Integrating generative AI into solar forecasting, SkyGPT \cite{nie2024skygpt} employs a physics-constrained stochastic video prediction model to synthesize future sky imagery. These generated sequences serve as high-fidelity inputs for a coupled U-Net architecture, effectively bridging generative atmospheric modeling with downstream photovoltaic power estimation.

\section{Proposed approach}\label{sec_work}

\subsection{Problem formulation}

The PV power generation forecast task (as depicted in Figure \ref{fig:framework}) can be mathematically written as a mapping function that learns from historical sky images and PV power outputs to predict future PV output. Following the works of \cite{nie2023skipp,nie2024skygpt}, we formulate the tasks as the function $\psi$.

\begin{equation}\label{eq:prediction}
\psi(I_{t_i-H:\delta:t_i}, P_{t_i-H:\delta:t_i}) =
\begin{cases}
P_{t_i+T} &\textrm{(single)}\\ 
P_{t_i+1:1:t_i+T} &\textrm{(multi)}
\end{cases}
\end{equation}

Here, $I$ and $P$ represent sky images and PV power output, respectively, where $i\in\mathbb{Z}:1\leq i\leq M$. $M$ is the total number of samples in the dataset. $H$ is the range of historical terms, $\delta$ is the interval between historical terms, and $T$ is the forecast horizon. In (\ref{eq:prediction}), the single-prediction task, denoted as a function $\psi$, takes images and PV power output of the past $H$ minutes to predict PV power output at $T$ minutes into the future. 

For the multi-horizon prediction task, the function takes the same inputs as the single-prediction task. However, in this task, we try to predict PV power output over the entire forecast horizon. Instead of a single prediction at time $T$, the multi-horizon task forecasts PV power output from time $t_i$ to $T$ at an interval of 1 min.

\subsection{Multi-Horizon approach for PV power output prediction}

We propose a novel approach for the PV power generation forecast task. The approach associates historical images and PV output with future PV output. Past works implement deep learning models to learn the relation between historical and future data. We present this association (or mapping) as two separate tasks --- single-point prediction and a multi-horizon approach. In single-point prediction, we aim to predict a single PV power output in the future. Sun \etal \cite{sun2018solar} implemented a convolutional neural network called SUNSET to predict the PV power generation 15 mins into the future. Nie \etal \cite{nie2023skipp} introduced a dataset of sky images and PV output named SKIPP'D (SKy Images and Photovoltaic Power Generation Dataset), where they used the SUNSET model to evaluate the efficacy of their dataset. Meanwhile, in SkyGPT \cite{nie2024skygpt}, a modified U-Net \cite{ronneberger2015u} is presented to predict the PV output at a 15-minute lead time.

While the above works focus on single-point prediction, we explore the multi-horizon approach for PV power output forecasting. This approach differs from single-point prediction in the sense that we jointly predict the PV power output for every minute leading up to the forecast horizon. In our experiments, we show that multi-horizon provides better accuracy than single-point prediction, which we attribute to the multi-horizon framework's ability to learn and model long-term temporal dependencies by jointly optimizing future PV output predictions. Moreover, the multi-horizon framework is architecture-independent and can be implemented with different baseline models to boost accuracy with minimal overhead computation.

\subsection{Baseline model}

We demonstrate the efficacy of our approach using two baseline models - SUNSET \cite{sun2018solar} and MobileNet (v3\_large) \cite{howard2019searching}. SUNSET proposed by Sun \etal is a CNN model for PV power generation prediction. Inspired by popular CNN networks like AlexNet, VGG, Inception, and ResNet, the authors adopted a similar Conv-Pool structure with several fully connected layers at the end in their SUNSET architecture. 

To further demonstrate the advantages of multi-horizon prediction across different architectures, we also train on the MobileNet architecture. We first pre-train the model on the ImageNet dataset \cite{deng2009imagenet}. After pretraining, we modify the network according to the requirements of our method. Since our approach takes multiple RGB images ($I_{t_i-H:\delta:t_i}$) as input, we replace the input (convolutional) layer with a convolutional layer that takes $H/\delta$ number of input features while the number of output features remains the same. We also freeze the fully-connected layers of the MobileNet. To process the historical PV power output ($P_{t_i-H:\delta:t_i}$), we pass it through a fully connected layer. Finally, we utilize the same fully-connected layers of the SUNSET model to process the image and PV power output features.

\begin{table}
\caption{Total number of training parameters (in millions) for each model when an input image size is $64\times64$. The multi-horizon prediction task achieves better performance than single-point prediction at a fractional increase in the number of training parameters.\label{tab:model_params}}
    \centering
    \begin{tabular}{c|cc}
        \hline
        Prediction & SUNSET & MobileNet\\
        \hline
        Single-point & 13.67 & 5.03\\
        Multi-horizon & 13.69 & 5.05\\
        \hline
    \end{tabular}
\end{table}

\subsection{Objective function}
We use the mean squared error (MSE) to train the network:
\begin{equation}\label{eq:loss}
\textrm{MSE}=\frac{1}{n}\sum^{n}_{i=1}(y_i-\hat{y}_i)^2,
\end{equation}
where $n$ is the number of samples, $\hat{y}_i$ is the predicted PV power output generated by the model, and $y_i$ is the ground-truth PV power output.

\section{Experimental Evaluation}\label{sec_results}
In this section, we evaluate the proposed approach using the baseline models - SUNSET and MobileNet. We take into account different image sizes, along with predictions at different times into future.

\subsection{Implementation details}
We conducted experiments using two baseline models, SUNSET and MobileNet, to evaluate the predictive performance of our approach against a single-horizon framework. Each model was trained using 10-fold cross-validation to ensure robustness and generalization. Training was performed for a maximum of 100 epochs with early stopping (patience of 5 epochs) to mitigate overfitting. We used the Adam optimizer with an initial learning rate of $3\times10^{-6}$, scheduled by cosine annealing to support smooth convergence. The input image resolutions of $64\times64$ and $224\times224$ were considered to analyze the effect of spatial resolution on performance. Predictions were made for future time points at $T$ = 15, 30, and 60 minutes for both models. Both models were implemented using PyTorch Lightning and trained on an NVIDIA Tesla V100 GPU. Table \ref{tab:model_params} summarizes the total number of trainable parameters for each model across both tasks. For the multi-horizon prediction task, the number of parameters increases only marginally—by 0.146\% for the SUNSET model and 0.398\% for MobileNet.

\subsection{Datasets}
\begin{figure*}[!t]
\centering
\includegraphics[scale=1.0]{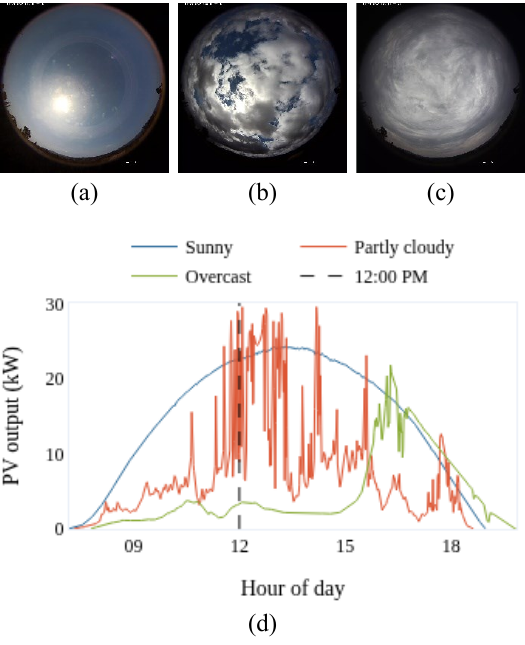}
\caption{Sky images taken at 12:00 PM on days with different weather conditions - (a) sunny (2017/09/15), (b) partly cloudy (2017/09/20), and (c) overcast (2017/08/04). (d) Minute-by-minute PV power output under these weather conditions. All images and PV power output are taken from the SKIPP'D dataset \cite{nie2023skipp}.}\label{fig:skippd}
\end{figure*}

We used the images and PV power output data from the SKIPP'D dataset for our approach. The dataset contains aligned pairs of sky images ($I$) and PV power output logs ($P$) from selected days of 2017 to 2019. Data is extracted at an interval of 1 minute during daytime (6:00 AM to 8:00 PM). The images obtained by a 6-megapixel 360-degree fish-eye camera have a resolution of $2048\times2048$ pixels. The entire dataset is split into training and test sets, each consisting of 349,372 and 14,003, respectively. The test set is manually selected to include 10 sunny and 10 cloudy days. Figure \ref{fig:skippd} illustrates the PV power output over three days under different weather conditions: sunny, partly cloudy, and overcast. It also presents the corresponding sky images captured at noon on each of these days.

\subsection{Qualitative and quantitative performances}\label{sec_results_qual_quant}
\begin{figure*}[!t]
\centering
\includegraphics[scale=0.9]{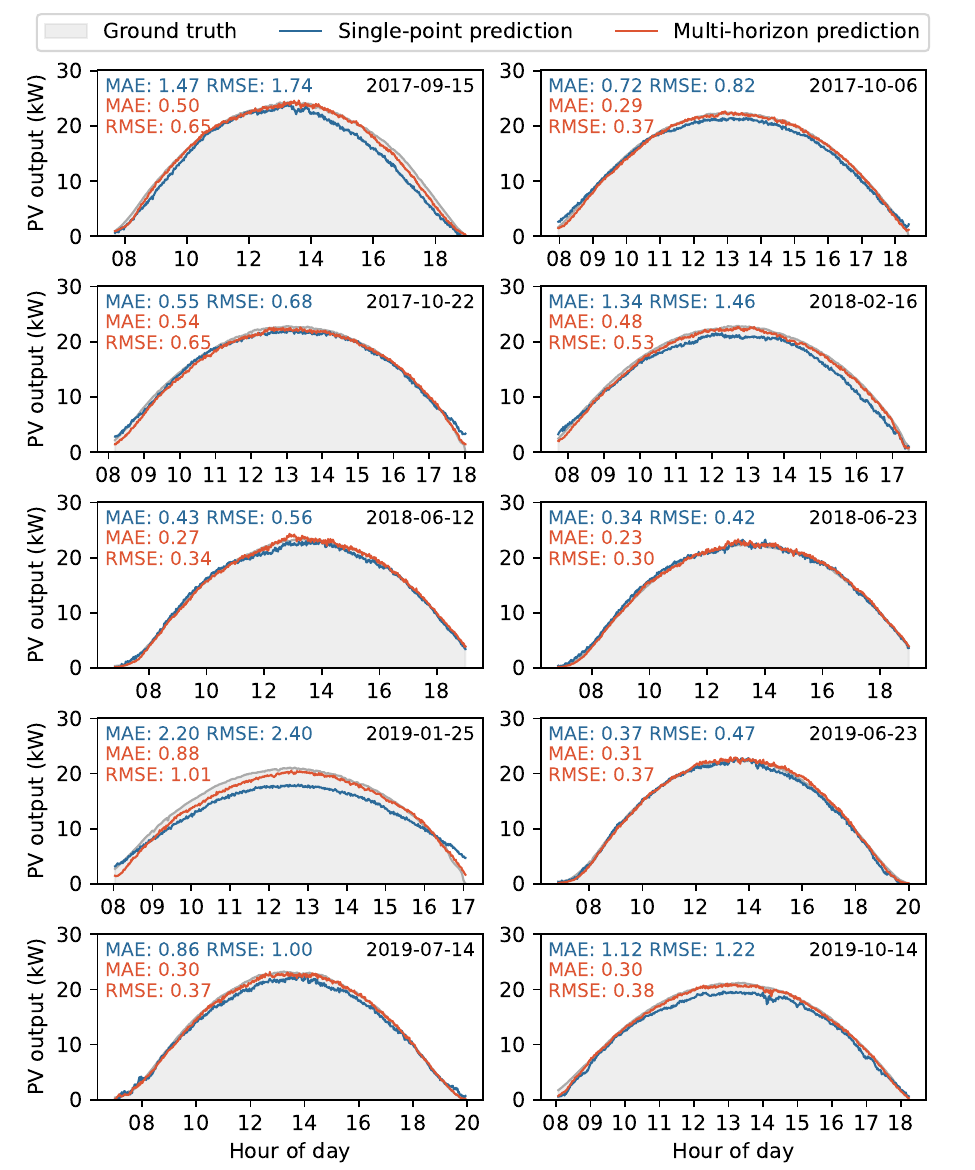}
\caption{Forecast results of the SUNSET model for 15-minute-ahead prediction on sunny days. The MAE and RMSE values for each task are displayed in the top-left corner of each plot for every day. Input image resolution of $64\times64$ is used.}\label{fig:sunset_sunny}
\end{figure*}

\begin{figure*}[!t]
\centering
\includegraphics[scale=0.9]{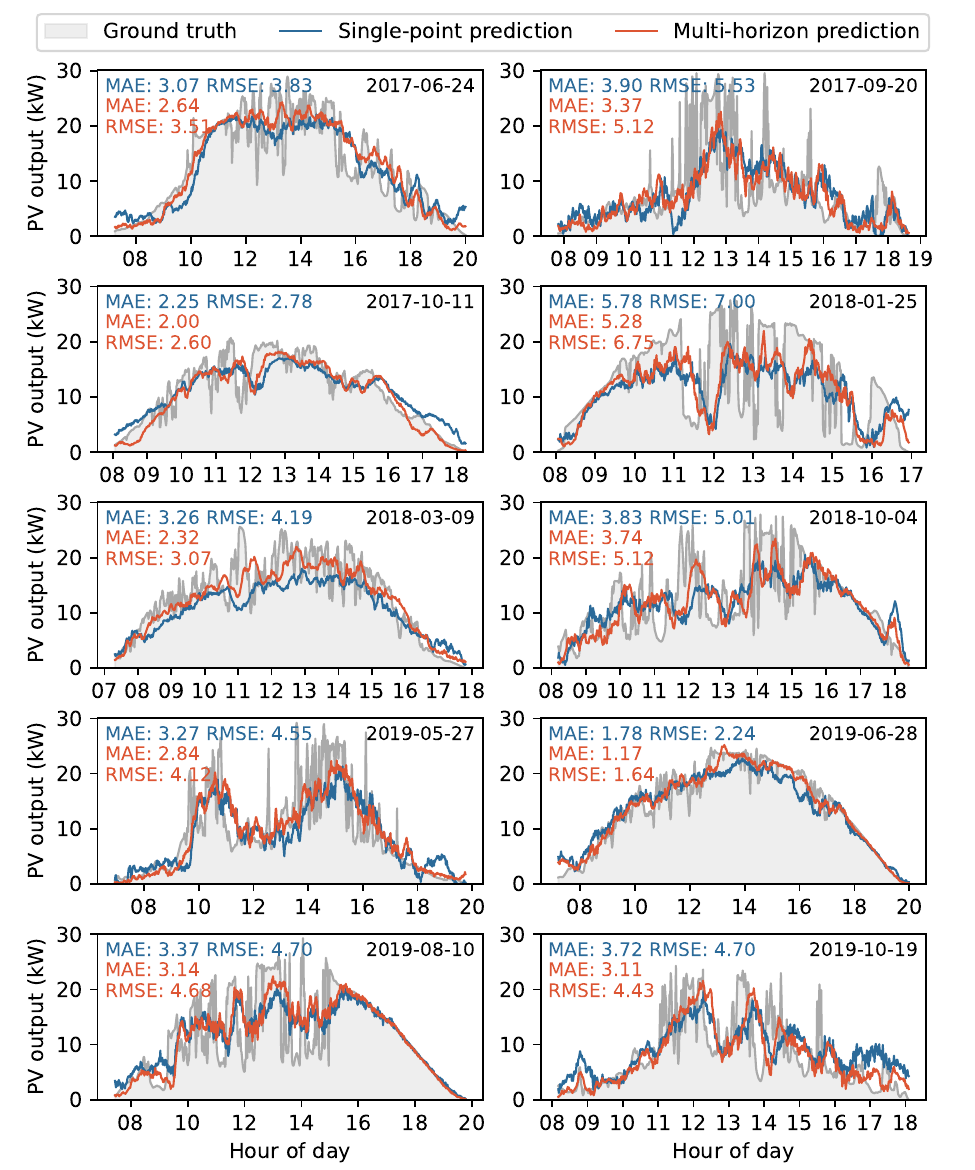}
\caption{Forecast results of the SUNSET model for 15-minute-ahead prediction on cloudy days. The MAE and RMSE values for each task are displayed in the top-left corner of each plot for every day. Input image resolution of $64\times64$ is used.}\label{fig:sunset_cloudy}
\end{figure*}

Figure \ref{fig:sunset_sunny} and \ref{fig:sunset_cloudy} depict the forecast prediction on the SUNSET models on the sunny and cloudy days, respectively, using $64\times64$ images. We provide a comparison of single and multi-horizon prediction tasks to predict the 15-minute future PV power output. We also measure the performance of the PV power output prediction by the following metrics: mean absolute error (MAE) and root mean squared error (RMSE). In both test days, we see an improvement from single-horizon prediction to multi-horizon prediction. On sunny and cloudy days, we see an improvement of $56.13\%$ and $6.67\%$ on RMSE scores, respectively. Overall on the entire test set, we see an improvement of $9.01\%$ in RMSE. The presence of large fluctuations on cloudy days diverts the model's prediction away from the ground truth and more towards the average values. Since the model relies on historical data, it follows a general trend and fails to anticipate random uncertainties in the future \cite{nie2023skipp}. This affects the scores on the cloudy test set and thereby on the entire test set. 


Table \ref{tab:results} shows the performance of both single-horizon and multi-horizon predictions using SUNSET and MobileNet architectures in terms of MAE and RMSE scores. To show that multi-horizon performs better with different input image sizes, we trained the models from scratch with input sizes of $64\times64$ and $224\times224$, while also considering prediction at different times in the future. All experiments are run using the same training hyperparameters, and the same 10-fold cross-validation approach is adopted. In all cases, we see fewer errors from multi-horizon prediction on the entire test dataset. On the SUNSET model with an input image size of $64\times64$, RMSE is reduced by 9\%, 5.6\%, and 3.4\% when predicting at 15, 30, and 60 mins ahead in the future, respectively.

\begin{table*}[!t]
\caption{Comparison of error metrics of single-point and multi-horizon prediction using SUNSET and MobileNet models with an input image size of $64\times64$ and $224\times224$. Predictions are generated for 15, 30, and 60 mins ahead in the future.\label{tab:results}}
\centering
\begin{tabular}{ccc|cc|cc|cc}
\hline
\multirow{2}{*}{Model} & \multirow{2}{*}{\begin{tabular}[c]{@{}c@{}}Pred.\\ (mins)\end{tabular}} & \multirow{2}{*}{Task} & \multicolumn{2}{c|}{Sunny days} & \multicolumn{2}{c|}{Cloudy days} & \multicolumn{2}{c}{Overall} \\ \cline{4-9} 
 &  &  & MAE & RMSE & MAE & RMSE & MAE & RMSE \\ \hline
\multirow{6}{*}{\begin{sideways}\begin{tabular}[c]{@{}c@{}}SUNSET\\ ($64\times64$)\end{tabular}\end{sideways}} & \multirow{2}{*}{15} & Single & 0.895 & 1.183 & 3.346 & 4.541 & 2.122 & 3.320 \\
 &  & Multi & 0.395 & 0.519 & 2.886 & 4.238 & 1.642 & 3.021 \\ \cline{2-9}
 & \multirow{2}{*}{30} & Single & 0.952 & 1.310 & 3.628 & 4.867 & 2.292 & 3.566 \\ 
 &  & Multi & 0.431 & 0.613 & 3.312 & 4.720 & 1.873 & 3.368 \\ \cline{2-9}
 & \multirow{2}{*}{60} & Single & 0.967 & 1.381 & 3.981 & 5.187 & 2.476 & 3.798 \\
 &  & Multi & 0.503 & 0.812 & 3.800 & 5.120 & 2.154 & 3.668 \\ \hline 
\multirow{6}{*}{\begin{sideways}\begin{tabular}[c]{@{}c@{}}SUNSET\\ ($224\times224$)\end{tabular}\end{sideways}} & \multirow{2}{*}{15} & Single & 1.034 & 1.251 & 3.483 & 4.661 & 2.260 & 3.414 \\
 &  & Multi & 0.561 & 0.717 & 3.056 & 4.274 & 1.810 & 3.066 \\ \cline{2-9} 
 & \multirow{2}{*}{30} & Single & 0.712 & 0.933 & 3.727 & 4.877 & 2.222 & 3.514 \\ 
 &  & Multi & 0.752 & 0.975 & 3.481 & 4.755 & 2.118 & 3.435 \\ \cline{2-9} 
 & \multirow{2}{*}{60} & Single & 0.935 & 1.156 & 4.059 & 5.210 & 2.499 & 3.776 \\ 
 &  & Multi & 0.946 & 1.320 & 3.945 & 5.124 & 2.448 & 3.744 \\ \hline 
\multirow{6}{*}{\begin{sideways}\begin{tabular}[c]{@{}c@{}}MobileNet\\ ($64\times64$)\end{tabular}\end{sideways}} & \multirow{2}{*}{15} & Single & 2.027 & 2.370 & 3.618 & 4.783 & 2.824 & 3.776 \\ 
 &  & Multi & 1.065 & 1.303 & 3.091 & 4.317 & 2.079 & 3.191 \\ \cline{2-9} 
 & \multirow{2}{*}{30} & Single & 2.161 & 2.468 & 3.726 & 5.004 & 2.945 & 3.947 \\ 
 &  & Multi & 0.941 & 1.239 & 3.395 & 4.722 & 2.169 & 3.454 \\ \cline{2-9} 
 & \multirow{2}{*}{60} & Single & 2.451 & 2.803 & 4.128 & 5.397 & 3.291 & 4.302 \\ 
 &  & Multi & 1.086 & 1.424 & 3.849 & 5.106 & 2.469 & 3.750 \\ \hline 
\multirow{6}{*}{\begin{sideways}\begin{tabular}[c]{@{}c@{}}MobileNet\\ ($224\times224$)\end{tabular}\end{sideways}} & \multirow{2}{*}{15} & Single & 1.157 & 1.440 & 3.244 & 4.519 & 2.202 & 3.355 \\
 &  & Multi & 0.367 & 0.506 & 3.067 & 4.603 & 1.719 & 3.276 \\ \cline{2-9} 
 & \multirow{2}{*}{30} & Single & 1.136 & 1.456 & 3.564 & 4.847 & 2.352 & 3.581 \\ 
 &  & Multi & 0.447 & 0.640 & 3.494 & 5.062 & 1.972 & 3.610 \\ \cline{2-9} 
 & \multirow{2}{*}{60} & Single & 1.378 & 1.725 & 3.911 & 5.089 & 2.646 & 3.802 \\ 
 &  & Multi & 0.522 & 0.747 & 3.762 & 5.102 & 2.144 & 3.649 \\ \hline 
\end{tabular}
\end{table*}

\subsection{Statistical analysis}
We also use the coefficient of determination, also known as $R^2$, to analyze the increased effectiveness of our approach. It measures the proportion of variance in the target variable that is explained by the predictive model. It provides an estimate of the model’s goodness of fit and its ability to generalize to unseen data by quantifying explained variance. It serves as an indicator of predictive performance, with a score of 1.0 denoting perfect predictions. $R^2$ is computed using the following equation -
\begin{equation}\label{eq:r2}
    R^2\{y,\hat{y}\}=1-\frac{\sum^{n}_{i=1}(y_i-\hat{y}_i)^2}{\sum^{n}_{i=1}(y_i-\bar{y})^2},
\end{equation}
where, $\bar{y}=\frac{1}{n}\sum^{n}_{i=1}y_i$ and $\sum^{n}_{i=1}(y_i-\hat{y}_i)^2=\sum^{n}_{i=1}\epsilon^{2}_{i}$. $\hat{y}_i$ is the predicted value of the $i$-th sample, $y_i$ is the corresponding true value for $n$ samples and $\epsilon_i$ denotes the residual for each sample.

\begin{figure*}[!t]
\centering
\includegraphics[scale=0.9]{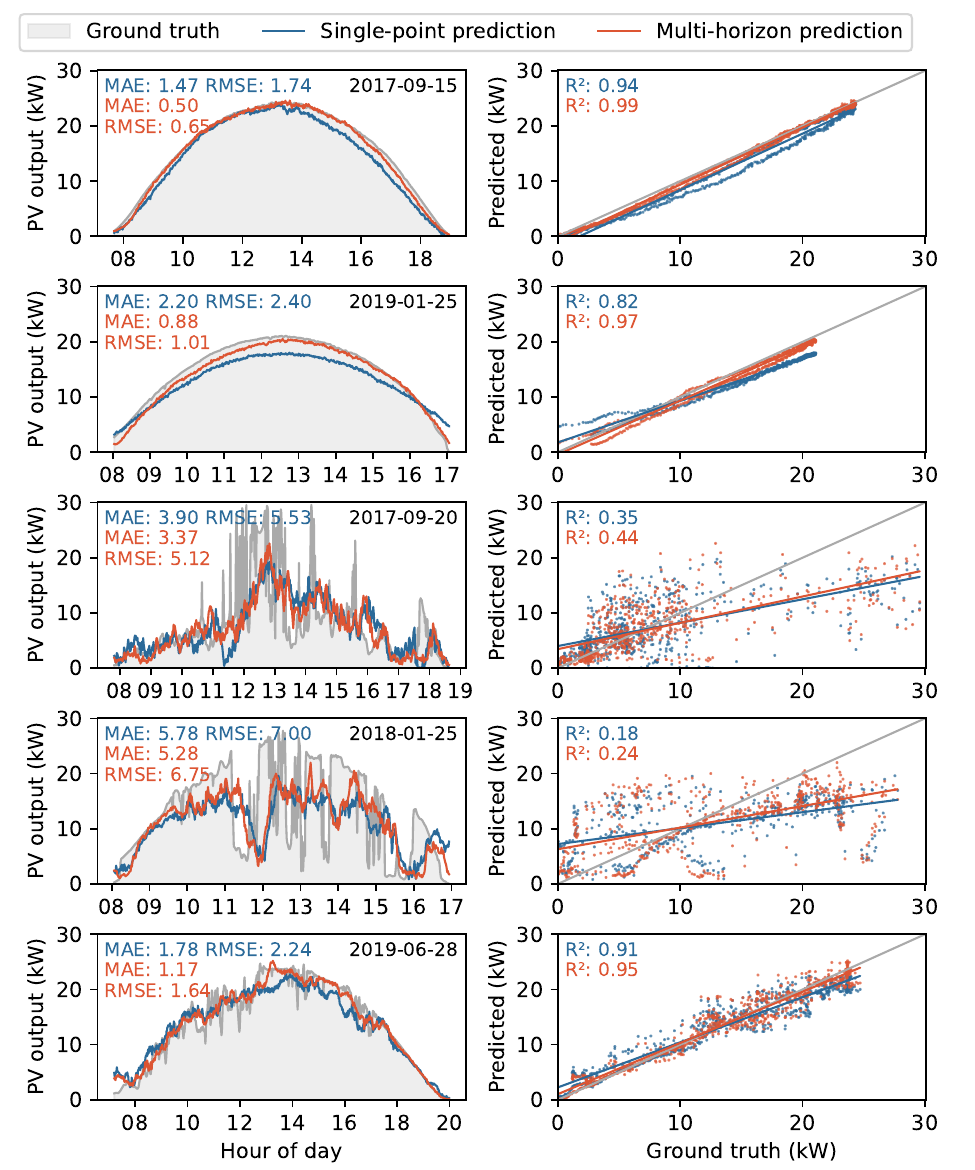}
\caption{The first column contains plots of selected days from Figures \ref{fig:sunset_sunny} and \ref{fig:sunset_cloudy}. The second column represents corresponding scatter plots, with their fitted least-squares regression line. The regression line corresponding to multi-horizon prediction shows closer alignment to the identity line (the 45-degree line) along with a better $R^2$ score, indicating improved prediction quality.}\label{fig:stat_result}
\end{figure*}

In Figure \ref{fig:stat_result}, we perform a statistical analysis of the two approaches on some selected days from the test set.  We analyze their prediction accuracy using scatter plots of ground-truth target values ($x$-axis) versus predicted outputs ($y$-axis). For each task, a least-squares regression line is fitted to the scatter plot to assess the linear agreement between predictions and true values. The regression line corresponding to multi-horizon prediction exhibits closer alignment to the identity line (i.e., the 45-degree line), indicating improved predictive fidelity relative to single-point prediction. This qualitative observation is corroborated by $R^2$, with the former yielding a higher score, signifying superior goodness of fit and enhanced generalization capability.

\subsection{Ablation study}
We also conduct an ablation study to evaluate the multi-horizon prediction performance with varying input images and PV data frequencies. In the experiments presented thus far, we have utilized the last 15 minutes ($H$) of historical images and PV power output, with a time interval ($\delta$) of 1 minute. In this study, we explore the effect of different time intervals ($\delta$) of 2 and 4 minutes on the prediction accuracy. Increasing the interval reduces the number of input images and corresponding PV data, i.e., with intervals of 1, 2, and 4 minutes, the number of input images and associated PV data are 16, 8, and 4, respectively.

\begin{figure*}[!t]
\centering
\includegraphics[scale=0.8]{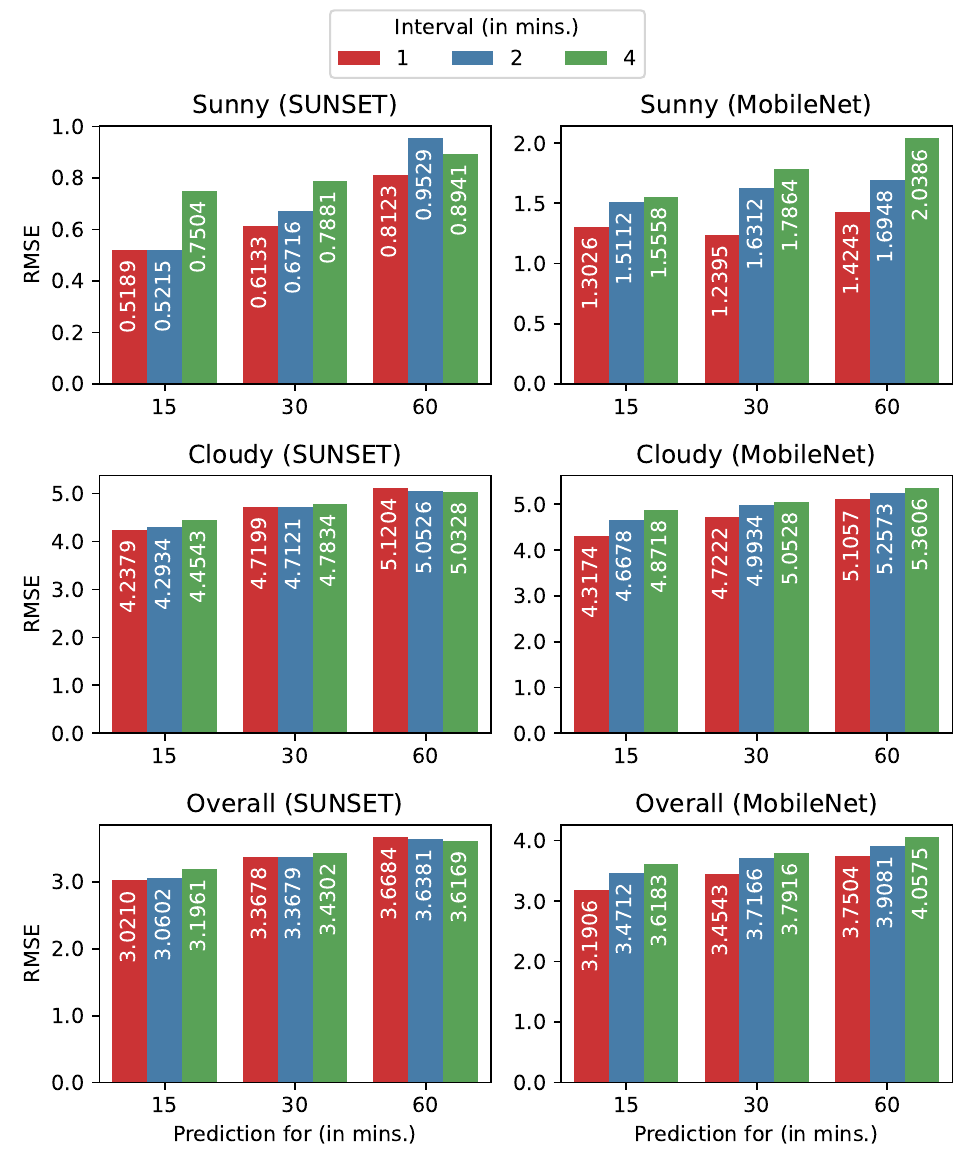}
\caption{Bar plot of average RMSE scores from multi-horizon prediction with prediction horizons ($T=15,30,60$ mins.). Each column corresponds to models - SUNSET and MobileNet. Plots in each row indicate outputs from different weather conditions. The intervals of 1, 2, and 4 minutes are indicated by red, blue, and green colors, respectively. Input image size of $64\times64$ is considered.}\label{fig:ch5_ablation}
\end{figure*}

Figure \ref{fig:ch5_ablation} presents a comparison of the average RMSE scores for the SUNSET and MobileNet models. Each row in the figure corresponds to different weather conditions: sunny, cloudy, and overall (both sunny and cloudy). The columns represent the two models — SUNSET and MobileNet. The x-axis indicates the prediction horizons at times $T = 15$, $T = 30$, and $T = 60$ minutes, while the y-axis shows the average RMSE values. The results for the intervals of 1, 2, and 4 minutes are indicated by red, blue, and green colors, respectively. A general trend is observed, where the RMSE score increases as the interval lengthens. This behavior is expected, as longer intervals provide fewer input images and less PV data, leading to reduced prediction accuracy. However, some exceptions are observed, particularly with the SUNSET model at the $T = 60$ minute prediction horizon.

\subsection{Analysis of weight gradients and filter diversity}
To better understand the effectiveness of the multi-horizon prediction performance over the single-point approach, we analyzed the weight gradients of convolutional layers of both models during training. The gradients calculate the derivative of the loss function with respect to each model parameter, indicating the direction and magnitude required to update weights for minimizing error. We notice that during the multi-horizon prediction task, the model takes more time to converge than the single-point prediction task. A short-term convergence can get the model stuck at a local minima and unable to estimate the optimal solution. A longer saturation phase enables the model to learn complex features, leading to robust performance on unseen data.

\begin{figure*}[!t]
\centering
\includegraphics[scale=0.8]{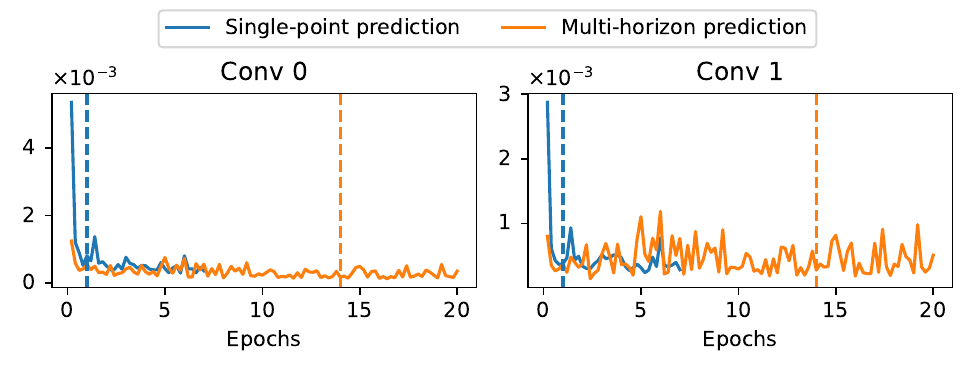}
\caption{Average absolute gradient magnitude for each layer of the SUNSET model across both tasks during training. The vertical dotted line indicates the epoch at which the minimum validation loss is attained, triggering early stopping.}\label{fig:ch5_grad_sunset}
\end{figure*}

\begin{figure*}[!t]
\centering
\includegraphics[scale=0.8]{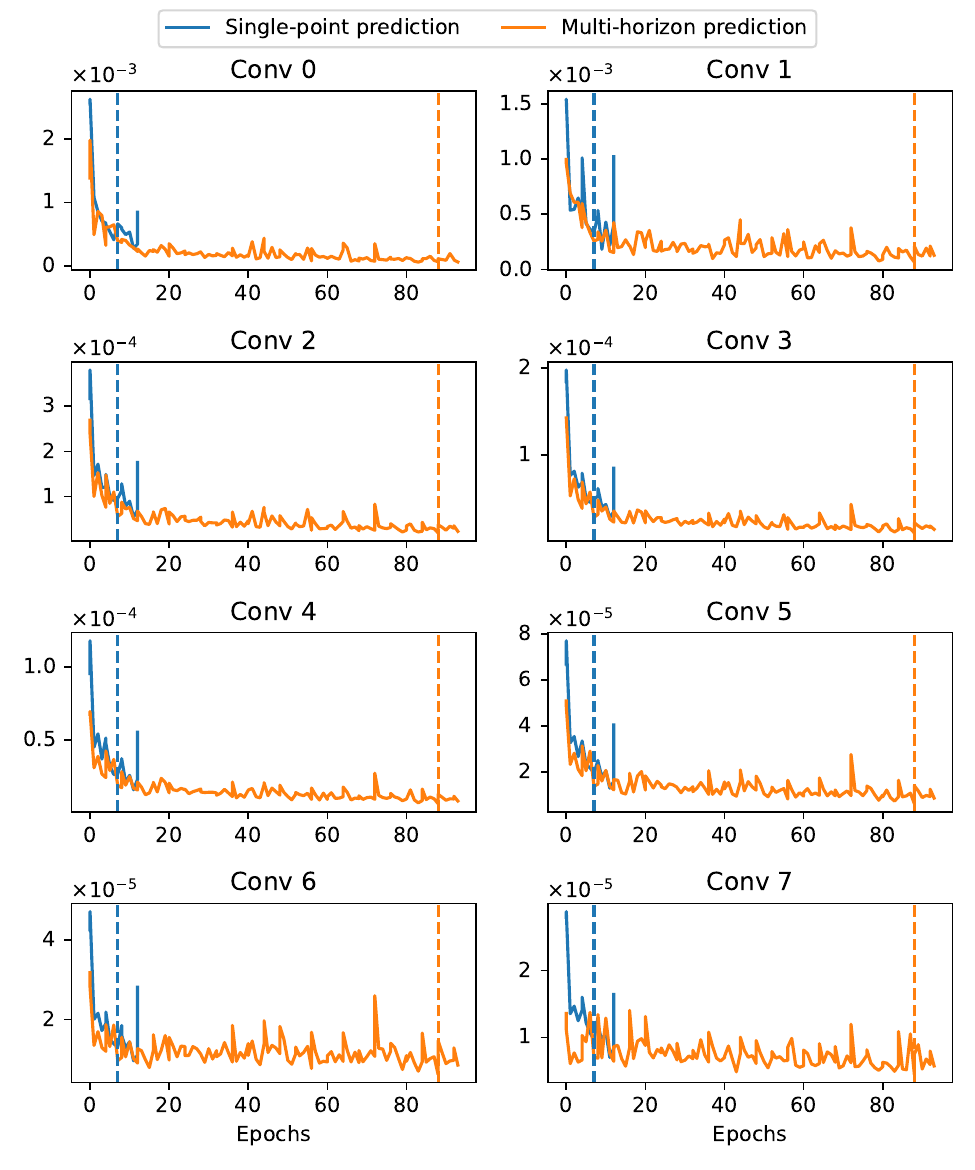}
\caption{Average absolute gradient magnitude for each layer of the MobileNet model across both tasks during training. The vertical dotted line indicates the epoch at which the minimum validation loss is attained, triggering early stopping.}\label{fig:ch5_grad_mobilenet}
\end{figure*}

We selected certain layers and computed the average of the absolute values of the gradients of those layer weights at a fixed train step interval. As the SUNSET model has only two convolutional layers, we monitor the gradients of both layers. However, given the depth of the MobileNet architecture and its numerous convolutional layers, we selected a representative subset of eight layers for analysis. Specifically, we included the first and last convolutional layers, as well as the first convolutional layer from each of the \nth{1}, \nth{3}, \nth{5}, \nth{8}, \nth{12}, and \nth{14} bottleneck blocks. Figure \ref{fig:ch5_grad_sunset} shows the mean absolute gradient values of the two convolutional layers of the SUNSET model during the training process. In both convolutional layers, the single-point prediction task leads to shorter convergence, while the multi-horizon prediction approach continues to learn features for a longer period, leading to better performance. Similarly, as depicted in Figure \ref{fig:ch5_grad_mobilenet}, we notice the same behavior for both tasks across the 8 selected layers when trained on the MobileNet model. 

\begin{figure*}[!t]
\centering
\includegraphics[scale=0.8]{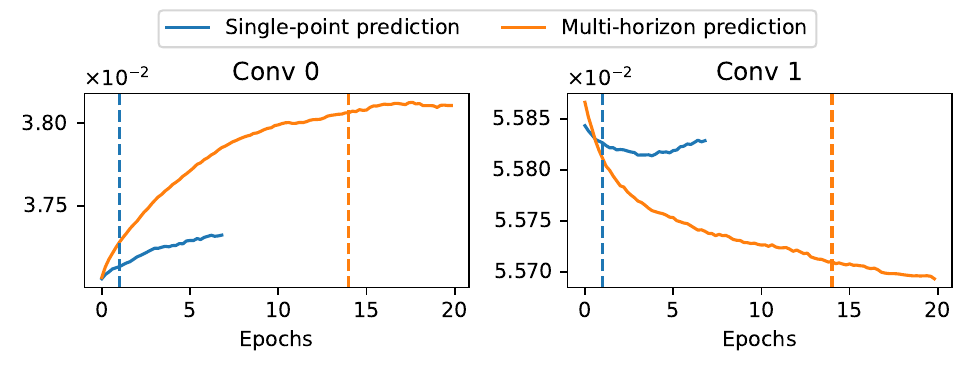}
\caption{Average absolute cosine similarity score for each layer of the SUNSET model across both tasks during training. The vertical dotted line indicates the epoch at which the minimum validation loss is attained, triggering early stopping.}\label{fig:ch5_sim_sunset}
\end{figure*}

\begin{figure*}
\centering
\includegraphics[scale=0.8]{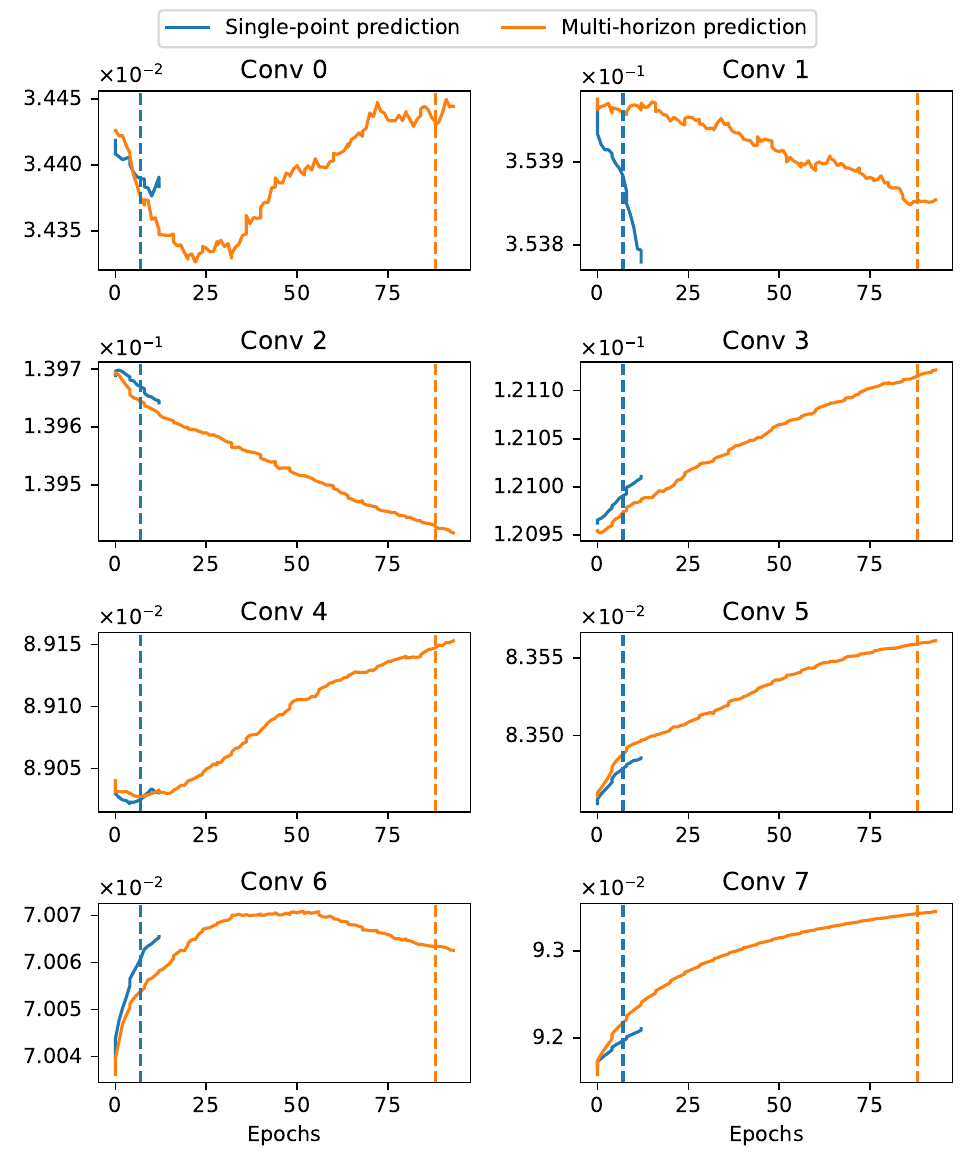}
\caption{Average absolute cosine similarity score for each layer of the MobileNet model across both tasks during training. The vertical dotted line indicates the epoch at which the minimum validation loss is attained, triggering early stopping.}\label{fig:ch5_sim_mobilenet}
\end{figure*}

Additionally, we explored the idea of filter diversity to analyze the performance of both tasks. Similar to filter (weight) gradients of the aforementioned layers, we estimated how diverse the filters are with one another in each layer in terms of cosine similarity score (presented in Equation \ref{eq:similarity}). The cosine similarity of vectors $x_1$ and $x_2$ is represented using a dot product and magnitude as
\begin{equation}\label{eq:similarity}
    \text{cosine similarity}=cos(\theta)=\frac{x_1 \cdot x_2}{{\left\| x_1\right\|} \cdot {\left\| x_2\right\|}}
\end{equation}
The similarity score ranges from -1 (exactly opposite) to +1 (exactly the same). However, in our computation, we took the absolute value of the score to remove redundant values. Figures \ref{fig:ch5_sim_sunset} and \ref{fig:ch5_sim_mobilenet} represent the average cosine similarity of the filters in each layer of the SUNSET and MobileNet models, respectively. Similar to gradient values, both tasks exhibit the same behavior based on the cosine similarity score. This validates the results presented in Section \ref{sec_results_qual_quant}.

\section{Discussion}\label{sec_discussion}
This work leverages existing baseline architectures, such as SUNSET and MobileNet, to predict future PV output by jointly optimizing intermediate PV output between the present and specified future prediction horizon. The proposed multi-horizon prediction can outperform single-point prediction tasks at a fractional increase in the number of training parameters (see Table \ref{tab:model_params} and Figures \ref{fig:sunset_sunny} and \ref{fig:sunset_cloudy}) in both SUNSET and MobileNet models. An improvement of 9\%, 5.6\%, and 3.4\% in terms of RMSE scores is noticeable (see Table \ref{tab:results}) on the SUNSET model when the prediction horizon was set at 15, 30, and 60 minutes ahead in the future, respectively.

The effectiveness of the proposed approach is also explained in terms of $R^2$ (see Figure \ref{fig:stat_result}), with a score of 1.0 denoting perfect predictions. In both sunny and cloudy days, the proposed method had higher $R^2$ values. In addition, we also fitted a least-squares regression line to assess the linear agreement between the predictions and the true values. The regression line corresponding to the multi-horizon prediction showed closer alignment to the identity line (the 45-degree line), indicating better predictive fit on that dataset. These findings are further supported by an analysis of filter gradients and filter diversity in selected convolutional layers during training. Figures \ref{fig:ch5_grad_sunset} and \ref{fig:ch5_grad_mobilenet} represent the mean absolute value of filter gradients of selected layers during the training process in the SUNSET and MobileNet models, respectively. In both cases, the multi-horizon prediction task exhibits a slower convergence rate, suggesting a more gradual optimization process that leads to improved generalization on unseen data. Additionally, the cosine similarity of the filters (see Figures \ref{fig:ch5_sim_sunset} and \ref{fig:ch5_sim_mobilenet}) also shows that both tasks exhibit the same behavior regardless of the underlying model architecture.

In future work, the following avenues can be explored to improve the performance of the approach. The models were trained and tested on the SKIPP'D dataset collected at Stanford University in California, United States. The weather conditions captured in these images are specific to the California climate. The models could be trained on datasets obtained from other parts of the world to incorporate diverse weather conditions. 

Additionally, the SKIPP'D dataset comprises PV measurements collected from a small-scale residential PV installation. In contrast, commercial and utility-scale PV systems often span large geographical areas. In such large-scale deployments, spatial variability in cloud cover can result in partial shading, where some panel arrays are affected by cloud shadows,  while others remain unshaded. Consequently, forecasting models intended for commercial or utility-scale applications may require training on those data that capture these spatial differences to achieve accurate PV power predictions.

In our experiments, we implemented the multi-horizon forecasting framework using baseline architectures such as SUNSET and MobileNet. With the rapid advancement of transformer-based and large language models, and their success across a wide range of computer vision tasks, extending these architectures to the present problem setting represents a promising research direction. Furthermore, network architecture selection and hyperparameter tuning may yield additional performance gains.

Cloud formation is governed by a complex interplay of atmospheric variables, including humidity, wind dynamics, temperature gradients, and related meteorological factors. A systematic analysis of these variables can provide deeper insights into cloud evolution and motion patterns. Consequently, incorporating these meteorological data into the forecasting framework has the potential to enhance predictive performance. Future work may pursue these research directions to further improve PV power forecasting accuracy.

\section{Conclusion}\label{sec_conclusion}
In this work, we propose a novel deep learning–based framework for PV power forecasting, termed multi-horizon prediction. The proposed method can leverage any baseline architectures, such as SUNSET and MobileNet, by jointly optimizing multiple horizons for PV output forecasts up to a specified prediction horizon. Comprehensive experiments are conducted across models with varying forecast horizons, which demonstrate a significant boost in accuracy compared with a single-horizon framework. The results consistently demonstrate that the proposed multi-horizon strategy outperforms conventional single-point prediction approaches, as evidenced by superior $R^2$ scores and overall predictive accuracy, which we attribute to the multi-horizon framework's ability to better learn and model long-term temporal dependencies in a time-series prediction through a joint optimization of multiple horizons.

\section*{Acknowledgements}
The funding for this project was provided by the Orlando Utilities Commission (OUC), Orlando, FL. The authors would like to thank Rubin York and Paul Brooker from the OUC for their helpful discussions and support.

\bibliographystyle{IEEEtran}
\bibliography{refs}

@article{sun2018solar,
  title={Solar PV output prediction from video streams using convolutional neural networks},
  author={Sun, Yuchi and Sz{\H{u}}cs, Gergely and Brandt, Adam R},
  journal={Energy \& Environmental Science},
  volume={11},
  number={7},
  pages={1811--1818},
  year={2018},
  publisher={Royal Society of Chemistry}
}

@article{MELLIT,
title = {A 24-h forecast of solar irradiance using artificial neural network: Application for performance prediction of a grid-connected PV plant at Trieste, Italy},
journal = {Solar Energy},
volume = {84},
number = {5},
pages = {807-821},
year = {2010},
issn = {0038-092X},
author = {Adel Mellit and Alessandro Massi Pavan},
}

@article{RANA,
title = {Univariate and multivariate methods for very short-term solar photovoltaic power forecasting},
journal = {Energy Conversion and Management},
volume = {121},
pages = {380-390},
year = {2016},
author = {Mashud Rana and Irena Koprinska and Vassilios G. Agelidis}
}

@article{SAHIN,
title = {Predictive modeling of PV solar power plant efficiency considering weather conditions: A comparative analysis of artificial neural networks and multiple linear regression},
journal = {Energy Reports},
volume = {10},
pages = {2837-2849},
year = {2023},
author = {Gökhan Sahin and Gültekin Isik and Wilfried G.J.H.M. {van Sark}},
}

@article{RUAN,
title = {On the use of sky images for intra-hour solar forecasting benchmarking: Comparison of indirect and direct approaches},
journal = {Solar Energy},
volume = {276},
pages = {112649},
year = {2024},
author = {Guoping Ruan and Xiaoyang Chen and Eng Gee Lim and Lurui Fang and Qi Su and Lin Jiang and Yang Du},
}

@article{FENG-Solarnet,
title = {SolarNet: A sky image-based deep convolutional neural network for intra-hour solar forecasting},
journal = {Solar Energy},
volume = {204},
pages = {71-78},
year = {2020},
author = {Cong Feng and Jie Zhang},
}

@article{SUN-Solar,
title = {Short-term solar power forecast with deep learning: Exploring optimal input and output configuration},
journal = {Solar Energy},
volume = {188},
pages = {730-741},
year = {2019},
author = {Yuchi Sun and Vignesh Venugopal and Adam R. Brandt},
}

@article{JONATHAN,
title = {A radiant shift: Attention-embedded CNNs for accurate solar irradiance forecasting and prediction from sky images},
journal = {Renewable Energy},
volume = {234},
pages = {121133},
year = {2024},
author = {Anto Leoba Jonathan and Dongsheng Cai and Chiagoziem C. Ukwuoma and Nkou Joseph Junior Nkou and Qi Huang and Olusola Bamisile},
}

@article{FU-indirect,
title = {Predicting solar irradiance with all-sky image features via regression},
journal = {Solar Energy},
volume = {97},
pages = {537-550},
year = {2013},
author = {Chia-Lin Fu and Hsu-Yung Cheng}
}

@article{MARQUEZ,
title = {Intra-hour DNI forecasting based on cloud tracking image analysis},
journal = {Solar Energy},
volume = {91},
pages = {327-336},
year = {2013},
author = {Ricardo Marquez and Carlos F.M. Coimbra}
}

@article{CHU-indirect,
title = {Hybrid intra-hour DNI forecasts with sky image processing enhanced by stochastic learning},
journal = {Solar Energy},
volume = {98},
pages = {592-603},
year = {2013},
author = {Yinghao Chu and Hugo T.C. Pedro and Carlos F.M. Coimbra}
}

@inproceedings{lu2025enhanced,
  title={Enhanced Solar Forecasting with Contrastive Learning Model: A 15-Minute Prediction},
  author={Lu, Chih-Yi and Chen, Ting-Yu and Hsieh, I-Yun Lisa},
  booktitle={2025 IEEE Industry Applications Society Annual Meeting (IAS)},
  pages={1--7},
  year={2025},
  organization={IEEE}
}

@inproceedings{leelaruji2020short,
  title={Short term prediction of solar irradiance fluctuation using image processing with resnet},
  author={Leelaruji, Thanonchai and Teerakawanich, Nithiphat},
  booktitle={2020 8th international electrical engineering congress (iEECON)},
  pages={1--4},
  year={2020},
  organization={IEEE}
}

@article{ZUO-indirect,
title = {Ten-minute prediction of solar irradiance based on cloud detection and a long short-term memory (LSTM) model},
journal = {Energy Reports},
volume = {8},
pages = {5146-5157},
year = {2022},
author = {Hui-Min Zuo and Jun Qiu and Ying-Hui Jia and Qi Wang and Fang-Fang Li},
}

@article{LIU,
title = {Spatial–temporal multimodal fusion model for intra-hour solar power forecasting under variable weather conditions},
journal = {Renewable Energy},
volume = {248},
pages = {123043},
year = {2025},
author = {Mengcheng Liu and Qiang Ling},
}

@article{ZANG,
title = {Improving ultra-short-term photovoltaic power forecasting using a novel sky-image-based framework considering spatial-temporal feature interaction},
journal = {Energy},
volume = {293},
pages = {130538},
year = {2024},
author = {Haixiang Zang and Dianhao Chen and Jingxuan Liu and Lilin Cheng and Guoqiang Sun and Zhinong Wei},
}

@ARTICLE{Taravat,
  author={Taravat, Alireza and Del Frate, Fabio and Cornaro, Cristina and Vergari, Stefania},
  journal={IEEE Geoscience and Remote Sensing Letters}, 
  title={Neural Networks and Support Vector Machine Algorithms for Automatic Cloud Classification of Whole-Sky Ground-Based Images}, 
  year={2015},
  volume={12},
  number={3},
  pages={666-670}}

@article{nie2023skipp,
  title={SKIPP’D: A SKy Images and Photovoltaic Power Generation Dataset for short-term solar forecasting},
  author={Nie, Yuhao and Li, Xiatong and Scott, Andea and Sun, Yuchi and Venugopal, Vignesh and Brandt, Adam},
  journal={Solar Energy},
  volume={255},
  pages={171--179},
  year={2023},
  publisher={Elsevier}
}

@article{nie2024skygpt,
  title={Skygpt: Probabilistic ultra-short-term solar forecasting using synthetic sky images from physics-constrained videogpt},
  author={Nie, Yuhao and Zelikman, Eric and Scott, Andea and Paletta, Quentin and Brandt, Adam},
  journal={Advances in Applied Energy},
  volume={14},
  pages={100172},
  year={2024},
  publisher={Elsevier}
}

@inproceedings{ronneberger2015u,
  title={U-net: Convolutional networks for biomedical image segmentation},
  author={Ronneberger, Olaf and Fischer, Philipp and Brox, Thomas},
  booktitle={International Conference on Medical image computing and computer-assisted intervention},
  pages={234--241},
  year={2015},
  organization={Springer}
}

@inproceedings{howard2019searching,
  title={Searching for mobilenetv3},
  author={Howard, Andrew and Sandler, Mark and Chu, Grace and Chen, Liang-Chieh and Chen, Bo and Tan, Mingxing and Wang, Weijun and Zhu, Yukun and Pang, Ruoming and Vasudevan, Vijay and others},
  booktitle={Proceedings of the IEEE/CVF international conference on computer vision},
  pages={1314--1324},
  year={2019}
}

@inproceedings{deng2009imagenet,
  title={Imagenet: A large-scale hierarchical image database},
  author={Deng, Jia and Dong, Wei and Socher, Richard and Li, Li-Jia and Li, Kai and Fei-Fei, Li},
  booktitle={2009 IEEE conference on computer vision and pattern recognition},
  pages={248--255},
  year={2009},
  organization={Ieee}
}

@article{barbieri2017very,
  title={Very short-term photovoltaic power forecasting with cloud modeling: A review},
  author={Barbieri, Florian and Rajakaruna, Sumedha and Ghosh, Arindam},
  journal={Renewable and Sustainable Energy Reviews},
  volume={75},
  pages={242--263},
  year={2017},
  publisher={Elsevier}
}

@article{sun2019short,
  title={Short-term solar power forecast with deep learning: Exploring optimal input and output configuration},
  author={Sun, Yuchi and Venugopal, Vignesh and Brandt, Adam R},
  journal={Solar Energy},
  volume={188},
  pages={730--741},
  year={2019},
  publisher={Elsevier}
}

@article{long2006retrieving,
  title={Retrieving cloud characteristics from ground-based daytime color all-sky images},
  author={Long, Charles N and Sabburg, Jeff M and Calb{\'o}, Josep and Pag{\`e}s, David},
  journal={Journal of atmospheric and oceanic technology},
  volume={23},
  number={5},
  pages={633--652},
  year={2006}
}

@article{ghonima2012method,
  title={A method for cloud detection and opacity classification based on ground based sky imagery},
  author={Ghonima, MS and Urquhart, B and Chow, CW and Shields, JE and Cazorla, Alberto and Kleissl, Jan},
  journal={Atmospheric Measurement Techniques},
  volume={5},
  number={11},
  pages={2881--2892},
  year={2012},
  publisher={Copernicus Publications G{\"o}ttingen, Germany}
}

@article{willert1991digital,
  title={Digital particle image velocimetry},
  author={Willert, Christian E and Gharib, Morteza},
  journal={Experiments in fluids},
  volume={10},
  number={4},
  pages={181--193},
  year={1991},
  publisher={Springer}
}

@article{beauchemin1995computation,
  title={The computation of optical flow},
  author={Beauchemin, Steven S. and Barron, John L.},
  journal={ACM computing surveys (CSUR)},
  volume={27},
  number={3},
  pages={433--466},
  year={1995},
  publisher={ACM New York, NY, USA}
}

@article{gielen2019role,
  title={The role of renewable energy in the global energy transformation},
  author={Gielen, Dolf and Boshell, Francisco and Saygin, Deger and Bazilian, Morgan D and Wagner, Nicholas and Gorini, Ricardo},
  journal={Energy strategy reviews},
  volume={24},
  pages={38--50},
  year={2019},
  publisher={Elsevier}
}

@article{kabir2018solar,
  title={Solar energy: Potential and future prospects},
  author={Kabir, Ehsanul and Kumar, Pawan and Kumar, Sandeep and Adelodun, Adedeji A and Kim, Ki-Hyun},
  journal={Renewable and Sustainable Energy Reviews},
  volume={82},
  pages={894--900},
  year={2018},
  publisher={Elsevier}
}

@article{chow2011intra,
  title={Intra-hour forecasting with a total sky imager at the UC San Diego solar energy testbed},
  author={Chow, Chi Wai and Urquhart, Bryan and Lave, Matthew and Dominguez, Anthony and Kleissl, Jan and Shields, Janet and Washom, Byron},
  journal={Solar Energy},
  volume={85},
  number={11},
  pages={2881--2893},
  year={2011},
  publisher={Elsevier}
}

@article{marquez2013intra,
  title={Intra-hour DNI forecasting based on cloud tracking image analysis},
  author={Marquez, Ricardo and Coimbra, Carlos FM},
  journal={Solar Energy},
  volume={91},
  pages={327--336},
  year={2013},
  publisher={Elsevier}
}

@article{quesada2014cloud,
  title={Cloud-tracking methodology for intra-hour DNI forecasting},
  author={Quesada-Ruiz, S and Chu, Y and Tovar-Pescador, J and Pedro, HTC and Coimbra, CFM},
  journal={Solar Energy},
  volume={102},
  pages={267--275},
  year={2014},
  publisher={Elsevier}
}

@article{chu2013hybrid,
  title={Hybrid intra-hour DNI forecasts with sky image processing enhanced by stochastic learning},
  author={Chu, Yinghao and Pedro, Hugo TC and Coimbra, Carlos FM},
  journal={Solar Energy},
  volume={98},
  pages={592--603},
  year={2013},
  publisher={Elsevier}
}

@article{chu2015real,
  title={Real-time prediction intervals for intra-hour DNI forecasts},
  author={Chu, Yinghao and Li, Mengying and Pedro, Hugo TC and Coimbra, Carlos FM},
  journal={Renewable energy},
  volume={83},
  pages={234--244},
  year={2015},
  publisher={Elsevier}
}

@article{chu2015short,
  title={Short-term reforecasting of power output from a 48 MWe solar PV plant},
  author={Chu, Yinghao and Urquhart, Bryan and Gohari, Seyyed MI and Pedro, Hugo TC and Kleissl, Jan and Coimbra, Carlos FM},
  journal={Solar Energy},
  volume={112},
  pages={68--77},
  year={2015},
  publisher={Elsevier}
}

@article{pedro2019adaptive,
  title={Adaptive image features for intra-hour solar forecasts},
  author={Pedro, Hugo TC and Coimbra, Carlos FM and Lauret, Philippe},
  journal={Journal of Renewable and Sustainable Energy},
  volume={11},
  number={3},
  year={2019},
  publisher={AIP Publishing}
}

@article{nie2020pv,
  title={PV power output prediction from sky images using convolutional neural network: The comparison of sky-condition-specific sub-models and an end-to-end model},
  author={Nie, Yuhao and Sun, Yuchi and Chen, Yuanlei and Orsini, Rachel and Brandt, Adam},
  journal={Journal of Renewable and Sustainable Energy},
  volume={12},
  number={4},
  year={2020},
  publisher={AIP Publishing}
}

@article{feng2020solarnet,
  title={SolarNet: A sky image-based deep convolutional neural network for intra-hour solar forecasting},
  author={Feng, Cong and Zhang, Jie},
  journal={Solar Energy},
  volume={204},
  pages={71--78},
  year={2020},
  publisher={Elsevier}
}

@article{paletta2020convolutional,
  title={Convolutional neural networks applied to sky images for short-term solar irradiance forecasting},
  author={Paletta, Quentin and Lasenby, Joan},
  journal={arXiv preprint arXiv:2005.11246},
  year={2020}
}

@article{nie2021resampling,
  title={Resampling and data augmentation for short-term PV output prediction based on an imbalanced sky images dataset using convolutional neural networks},
  author={Nie, Yuhao and Zamzam, Ahmed S and Brandt, Adam},
  journal={Solar Energy},
  volume={224},
  pages={341--354},
  year={2021},
  publisher={Elsevier}
}

@article{feng2022convolutional,
  title={Convolutional neural networks for intra-hour solar forecasting based on sky image sequences},
  author={Feng, Cong and Zhang, Jie and Zhang, Wenqi and Hodge, Bri-Mathias},
  journal={Applied Energy},
  volume={310},
  pages={118438},
  year={2022},
  publisher={Elsevier}
}

\end{document}